\documentclass{article}
\usepackage{spconf,amsmath,epsfig}
\usepackage{xcolor}
\usepackage{enumitem}
\usepackage{graphicx}
\usepackage{amsmath}
\usepackage{amssymb}
\usepackage{epsfig}
\usepackage{graphicx}
\usepackage{caption}
\usepackage{adjustbox}
\usepackage[utf8]{inputenc}
\usepackage[T1]{fontenc}
\usepackage{threeparttable}
\usepackage{bbm}
\usepackage{booktabs}
\usepackage{multirow}
\usepackage{colortbl}
\usepackage{comment}
\definecolor{cvprblue}{rgb}{0.21,0.49,0.74}
\usepackage[pagebackref,breaklinks,colorlinks,citecolor=cvprblue]{hyperref}
\usepackage{pgfplots}
\usepackage{pgfplotstable}
\usepackage{hyperref}
\usepackage{textcomp}

\newcommand{\Th}[1]{\textsc{#1}}

\newcommand{\red}[1]{{\textcolor{red}{#1}}}

\newcommand{\citeme}[1]{\red{[XX]}}
\newcommand{\refme}[1]{\red{(XX)}}

\makeatletter
\newcommand*\bdot{\mathpalette\bdot@{.7}}
\newcommand*\bdot@[2]{\mathbin{\vcenter{\hbox{\scalebox{#2}{$\m@th#1\bullet$}}}}}
\makeatother

\makeatletter
\DeclareRobustCommand\onedot{\futurelet\@let@token\@onedot}
\def\@onedot{\ifx\@let@token.\else.\null\fi\xspace}

\makeatother

\definecolor{TableColor}{rgb}{0.835, 0.894, 0.968}
\definecolor{CommonChoices}{rgb}{0.935, 0.935, 0.935}

\newcommand{\Col}[2]{{\color{#1}#2}}

\newcommand{\bill}[1]{\Col{purple}{#1}}

\newcommand{\gain}[1]{}

\definecolor{YOLOv8nano}{RGB}{0, 0, 255} 
\definecolor{YOLOv8small}{RGB}{255, 182, 193} 
\definecolor{YOLOv8medium}{RGB}{255, 222, 173} 
\definecolor{YOLOv8large}{RGB}{128, 128, 128} 

\title{Evaluation of resource-efficient Object Detectors for Crater Detection on Embedded Systems} 

\title{Evaluation of Embedded System Solutions for Crater Detection 
onboard satellites with resource-efficient AI }
\title{Evaluation of Embedded System Solutions for Crater Detection 
with computationally-demanding AI}
\title{Evaluation of Embedded Solutions for Crater Detection 
with computationally-demanding AI}
\title{Evaluation of resource-efficient crater Detectors  on Embedded Systems}

\authors
  {Simon Vellas$^{1}$, Bill Psomas$^{1}$, Kalliopi Karadima$^{1}$, Dimitrios Danopoulos$^{1}$,  \\ Alexandros Paterakis$^{1}$, George Lentaris$^{1}$, Dimitrios Soudris$^{1}$, Konstantinos Karantzalos$^{1}$\vspace{0.5em}} 
  {$^1$National Technical University of Athens}
  
\begin{document}
%

\makeatletter
\apptocmd\@maketitle{{\teaser{}}}{}{}
\makeatother
\newcommand{\myfig}[1]{\includegraphics[width=0.10\textwidth]{#1}}
\newcommand{\mycaption}[1]{{\rotatebox[origin=c]{0}{\footnotesize#1}}}

\newcommand{\teaser}{%
\vspace{-15pt}
\centering
\setlength{\tabcolsep}{2pt}
\begin{tabular}{ccccccccc}

    \myfig{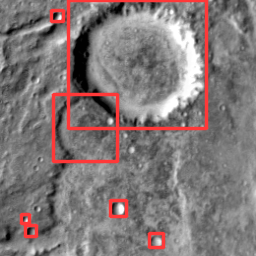} &
    \myfig{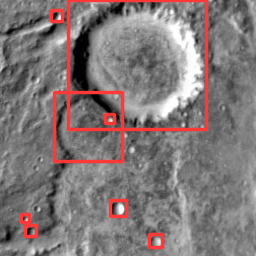} &
    \myfig{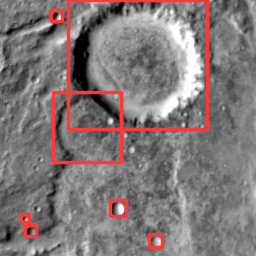} &
    \myfig{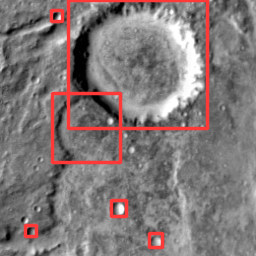} &
    \myfig{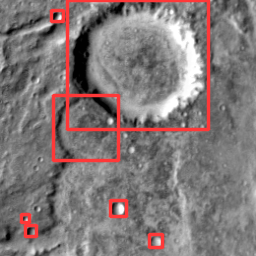} &
    \myfig{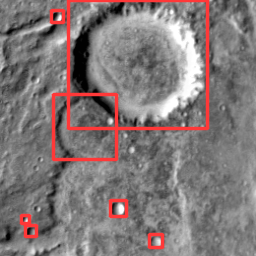} &
    \myfig{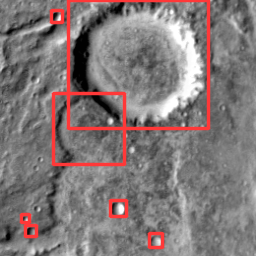} &
    \myfig{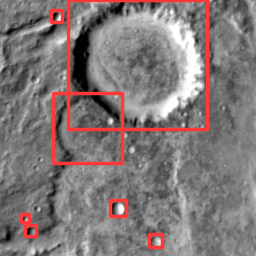} &
    \myfig{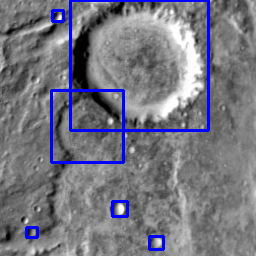} \\


    \vspace{-3pt}

    \mycaption{YOLOv5}&
    \mycaption{YOLOv5}&
    \mycaption{YOLOv5}&
    \mycaption{YOLOv5}&
    \mycaption{YOLOv8}&
    \mycaption{YOLOv8}&
    \mycaption{YOLOv8}&
    \mycaption{YOLOv8}& 
    \mycaption{ground}\\

    \mycaption{nano}&
    \mycaption{small}&
    \mycaption{medium}&
    \mycaption{large}&
    \mycaption{nano}&
    \mycaption{small}&
    \mycaption{medium}&
    \mycaption{large} &
    \mycaption{truth}\\

\end{tabular}
\vspace{-6pt}
\captionof{figure}{\emph{Qualitative comparison of crater detection results} of YOLOv5~\cite{yolov5} and YOLOv8~\cite{yolov8} on a Mars image~\cite{aichallenge}. Visualizations show the \textcolor{red}{predicted} bounding boxes across different model sizes. The final column shows the \textcolor{blue}{groundtruth} for reference.}

\label{fig:teaser}
\par\vspace{10pt}
}

\maketitle
\begin{abstract}
Real-time analysis of Martian craters is crucial for mission-critical operations, including safe landings and geological exploration. This work leverages the latest breakthroughs for on-the-edge crater detection aboard spacecraft. We rigorously benchmark several YOLO networks using a Mars craters dataset, analyzing their performance on embedded systems with a focus on optimization for low-power devices. We optimize this process for a new wave of cost-effective, commercial-off-the-shelf-based smaller satellites. Implementations on diverse platforms, including Google Coral Edge TPU, AMD Versal SoC VCK190, Nvidia Jetson Nano and Jetson AGX Orin, undergo a detailed trade-off analysis. Our findings identify optimal network-device pairings, enhancing the feasibility of crater detection on resource-constrained hardware and setting a new precedent for efficient and resilient extraterrestrial imaging. Code at: \href{https://github.com/billpsomas/mars_crater_detection}{https://github.com/billpsomas/mars\_crater\_detection}.


\end{abstract}
\vspace{-5pt}
\section{Introduction}
\label{sec:intro}

\emph{Impact craters} are prevalent across the solar system, including planets, moons, asteroids, transneptunian objects, and occasionally comets~\cite{craters_on_comets}. They form from meteoroid, asteroid, or comet impacts~\cite{crater_creation_mars&moon, lunar_crater_creation}, and are critical for understanding \emph{geomorphological processes}~\cite{mars_insights}.

Crater \emph{studies} offer insights into the early evolution of solar system bodies, such as moons orbiting planets~\cite{formation_phobos, pluto_charon}. They help determine the ages of planetary bodies~\cite{aging_Hartmann, aging_Neumann, aging_Michael, lunar_aging}, study material deposition, and explore climatic history and potential habitability~\cite{Barlow_geology, Barlow_subsurface, Hartmann_geology, mars_moon_morphology, mars_climate, climate_Gale_crater}.


Crater \emph{detection} is vital for autonomous navigation and landing in space missions. Despite advances in deep learning, there is a lack of thorough evaluation for \emph{real-world}, \emph{onboard space applications}. Automated crater detection can improve navigation accuracy and safety in space missions~\cite{spacecraft_navigation}, particularly for autonomous probes and landers~\cite{autonomous_operation, Landing_ESA, planetary_landing}.


Initially, crater detection relied on \emph{manual inspection} by experts, resulting in extensive crater databases for Mars and the Moon~\cite{Robbins_database, THEMIS_2004, lunar_database}. This approach was time-consuming, labor-intensive, and prone to variability~\cite{Robbins_variability}, leading to the development of automated \emph{crater detection algorithms} utilizing features like shape, edge, and contour~\cite{in_situ_Mars, in_situ_Mars.2, CDA_Mars_2005, CDA_Moon, CDA_Mars_2012, COSMIC}. However, these algorithms faced challenges in large-scale and diverse diameter range detection~\cite{CDA_comparisons, CDA_overview, small_craters_IEEE, Catalogues}.


Deep learning has enabled more accurate and efficient crater detection~\cite{DL, small_scale_DL, tewari2023deep}. Deploying these algorithms in space missions poses challenges regarding \emph{reliability} in harsh environments. CMOS technologies are prone to space-induced errors, necessitating \emph{fault-tolerant} techniques for commercial off-the-shelf processors and cameras. Issues like transient bit-flips in SRAM-based FPGAs or accumulated dose effects on camera sensors can impact performance~\cite{9758864, HUBER2017669}. Thus, optimizing networks for embedded systems and enhancing their \emph{resilience} against radiation effects through software mitigation is crucial.


This work focuses on training YOLO~\cite{yolo, yolov4, yolov8} networks for real-time crater detection (\autoref{fig:teaser}). YOLO's object detection capabilities and efficiency make it suitable for planetary exploration missions. Evaluating these networks on embedded systems and conducting network/device \emph{pair exploration} and \emph{trade-off analysis} aims to improve resilient deep learning solutions in orbit.


In summary, we make the following contributions:
\begin{enumerate}[itemsep=2pt, parsep=0pt, topsep=3pt]
    \item We train several efficient crater detection networks on a publicly available benchmark dataset.
    \item We evaluate these networks on several embedded processing systems.
    \item We conduct a network/device pair exploration and trade-off analysis.
\end{enumerate}
\vspace{-5pt}
\section{Related Work}
\label{sec:related}

\textbf{Crater Detection:} Methodologies have evolved from \emph{manual inspection} by experts to \emph{automated algorithms} and, more recently, \emph{deep learning}. Initially, planetary geologists manually inspected high-resolution images to identify and catalog craters~\cite{standards_1979}. This method, though foundational, was limited by \emph{subjective interpretation} and lacked scalability.

Automated \emph{crater detection algorithms} (CDA) emerged to address these challenges. Techniques include the combinational Hough transform~\cite{CHT_Honda}, continuously scalable template matching (CSTM)~\cite{template_matching}, shape recognition~\cite{shape_rec}, and processing highlight and shadow features using inverted images~\cite{bright_region}.

Detecting craters at large scales and across diverse diameters is challenging, but some studies have addressed these for global surveys~\cite{large_scale_DeLatte}. Deep learning has further enhanced crater detection by allowing algorithms to develop their own features, improving efficiency and accuracy~\cite{large_scale_Silburt, DL, small_scale_DL, DL_review}. Recent methods utilize convolutional neural networks to automatically learn and extract features, significantly advancing the field.

\textbf{Benchmarking:} Accurate \emph{benchmarking} of processing architectures ensures they meet the \emph{computational demands} of modern edge applications, such as data processing, scientific measurements, autonomous navigation, and real-time decision-making. Simulating operational environments and stress-testing capabilities help identify potential failures and \emph{optimize} configurations. Due to constant advancements, reviewing the state-of-the-art in embedded computing is essential~\cite{Naidoo2024}. The rapid growth of AI necessitates low-latency, efficient data processing close to the source~\cite{s23239495, woisetschläger2023fledge}.

Machine learning algorithms on spacecraft are critical for autonomous navigation and landing. The rise of commercially available embedded devices that address increased processing needs while maintaining low power consumption has garnered significant interest in the space community~\cite{glentar}. This interest is evident in mission-specific applications such as earth observation, remote sensing~\cite{furano}, and spacecraft pose estimation~\cite{aerospace7110159}.
\vspace{-5pt}
\section{Problem Formulation}
\label{sec:method}

Given an image \( X \in \mathbb{R}^{H \times W \times C} \), where \( H \) is the height, \( W \) is the width, and \( C = 1 \) for grayscale images, the task is to detect craters. The goal is to train an object detection model \( f \), parameterized by \( \theta \), that predicts bounding boxes and confidence scores for craters:
\begin{equation}
\hat{y} = f(X; \theta)
\end{equation}
The predictions \( \hat{y} \) include bounding boxes \( B \) and confidence scores \( S \):
\begin{equation*}
\hat{y} = \{(B_i, S_i) \mid i = 1, \ldots, N\},
\end{equation*}
where \( N \) is the number of predictions. Each \( B_i \) is represented by \( (x_i, y_i, w_i, h_i) \), with \( (x_i, y_i) \) as the center and \( w_i, h_i \) as the width and height. The confidence score \( S_i \in [0, 1] \) indicates the likelihood of a crater.

Let \( y \) be the ground truth annotations:
\begin{equation*}
y = \{B^{gt}_j \mid j = 1, \ldots, M\},
\end{equation*}
where \( M \) is the number of ground truth bounding boxes. The loss function \( L \) used to train \( f \) consists of localization loss \( L_{loc} \) and confidence loss \( L_{conf} \). The total loss \( L \) is:
\begin{equation}
L = \alpha L_{loc} + \beta L_{conf}
\end{equation}
The training objective is to find the optimal parameters \( \theta^* \) that minimize \( L \):
\begin{equation}
 \theta^* = \arg\min_\theta L(f(X; \theta), y)
\end{equation}
By optimizing this, the model \( f \) learns to detect craters accurately, providing bounding boxes and confidence scores.

\vspace{-5pt}
\section{Experiments}
\label{sec:exp}

\subsection{Models}
\label{sec:models}

\begin{table*}[ht]
\centering
\resizebox{0.8\textwidth}{!}{
\begin{tabular}{ccccccccccc}
\toprule
\multirow{2}{*}{\Th{Device}} & \multirow{2}{*}{\Th{Power}} & \multirow{2}{*}{\Th{Accuracy}} & \multicolumn{4}{c}{\Th{YOLO v5~\cite{yolov5}}} & \multicolumn{4}{c}{\Th{YOLO v8~\cite{yolov8}}} \\ \cmidrule{4-11}
& \Th{Consumption} & \& \Th{Latency} & \textcolor{YOLOv8nano}{\Th{nano}} & \textcolor{YOLOv8small}{\Th{small}} & \textcolor{YOLOv8medium}{\Th{medium}} & \textcolor{YOLOv8large}{\Th{large}} & \textcolor{YOLOv8nano}{\Th{nano}} & \textcolor{YOLOv8small}{\Th{small}} & \textcolor{YOLOv8medium}{\Th{medium}} & \textcolor{YOLOv8large}{\Th{large}}\\
\midrule
\multirow{4}{*}{Intel Xeon x86 CPU (1-core)} & \multirow{4}{*}{100W} & \Th{AP$_{30:60}$} & 79.6 & 78.0 & 80.8 & 81.2 & 77.9 & 77.7 & 80.7 & 81.9\\
                                        & & \Th{AP$_{50}$} & 80.2 & 78.0 & 81.4 & 80.9 & 78.9 & 78.3 & 80.1 & 82.5 \\
                                        & & \Th{AR$_{50}$} & 89.3 & 85.8 & 89.6 & 88.6 & 86.3 & 86.5 & 88.9 & 91.0\\ \cmidrule{3-11}
                                    & &{\emph{t} (msec):} & 30 & 72 & 175 & 340 & 33.0 & 82 & 204 & 397  \\
\midrule
\multirow{4}{*}{NVIDIA V100 Tensor Core GPU$^{1}$} & \multirow{4}{*}{300W} & \Th{AP$_{30:60}$} & 79.8 & 77.7 & 80.8 & 81.0 & 77.9 & 77.7 & 80.9 & 81.9\\
                                        & & \Th{AP$_{50}$} & 80.3 & 77.8 & 81.4 & 80.8 & 78.9 & 78.3 & 80.2 & 82.5 \\
                                        & & \Th{AR$_{50}$} & 89.4 & 85.8 & 89.6 & 88.5 & 86.3 & 86.3 & 89.0 & 91.0\\ \cmidrule{3-11}
                                      & & {\emph{t} (msec):} & 1.2 & 1.5 & 2.3 & 3.1 & 1.3 & 1.6 & 2.5 & 3.3 \\
\midrule
\multirow{4}{*}{ARM64 Cortex-A CPU (4-cores) \tnote{1}} & \multirow{4}{*}{5-15W} & \Th{AP$_{30:60}$} & 79.6 & 78.0 & 80.8 & 81.2 & 77.9 & 77.7 & 80.7 & 81.9\\
                                        & & \Th{AP$_{50}$} & 80.2 & 78.0 & 81.4 & 80.9 & 78.9 & 78.3 & 80.1 & 82.5 \\
                                        & & \Th{AR$_{50}$} & 89.3 & 85.8 & 89.6 & 88.6 & 86.3 & 86.5 & 88.9 & 91.0\\ \cmidrule{3-11}
                                      & & {\emph{t} (msec):} & 104.3 & 214.5 & 478.5 & 850.3 & 114.8 & 246.9 & 538.4 & 990.1 \\
\midrule
\multirow{4}{*}{Google Coral edge TPU 1GB (INT8) \tnote{1}} & \multirow{4}{*}{5W} & \Th{AP$_{30:60}$} & 72.5 & 73.5 & 73.7 & 75.4 & 70.3 & 73.2 & 77.8 & 77.2 \\
                                        & & \Th{AP$_{50}$} & 72.4 & 74.8 & 73.4 & 76.2 & 70.1 & 72.9 & 77.8 & 78.5 \\
                                        & & \Th{AR$_{50}$} & 82.1 & 82.6 & 83.5 & 84.2 & 78.0 & 81.4 & 87.9 & 87.9\\ \cmidrule{3-11}
                                      & & {\emph{t} (msec):} & 17.3 & 26.3 & 69.5 & 135.6 & 18.3 & 32.7 & 73.5 & 124.1  \\
\midrule
\multirow{4}{*}{NVIDIA Jetson Nano 4GB \tnote{1}} & \multirow{4}{*}{10W} & \Th{AP$_{30:60}$} & 79.6 & 78.0 & 80.8 & 81.2 & 77.9 & 77.7 & 80.7 & 81.9\\
                                        & & \Th{AP$_{50}$} & 80.2 & 78.0 & 81.4 & 80.9 & 78.9 & 78.3 & 80.1 & 82.5 \\
                                        & & \Th{AR$_{50}$} & 89.3 & 85.8 & 89.6 & 88.6 & 86.3 & 86.5 & 88.9 & 91.0  \\ \cmidrule{3-11}
                                      & & {\emph{t} (msec):} & 24.0 & 35.2 & 75.5 & 151.0 & 25.7 & 40.5 & 79.4  & 159.1  \\
\midrule
\multirow{4}{*}{NVIDIA Jetson AGX Orin 64GB \tnote{1}} & \multirow{4}{*}{15W mode} & \Th{AP$_{30:60}$} & 79.6 & 78.0 & 80.8 & 81.2 & 77.9 & 77.7 & 80.7 & 81.9\\
                                        & & \Th{AP$_{50}$} & 80.2 & 78.0 & 81.4 & 80.9 & 78.9 & 78.3 & 80.1 & 82.5 \\
                                        & & \Th{AR$_{50}$} & 89.3 & 85.8 & 89.6 & 88.6 & 86.3 & 86.5 & 88.9 & 91.0\\ \cmidrule{3-11}
                                      & & {\emph{t} (msec):} & 10.7 & 14.4 & 22.9 & 34.1 & 9.5 & 15.4 & 23.9 & 38.0 \\
\midrule
\multirow{4}{*}{NVIDIA Jetson AGX Orin 64GB \tnote{1}} & \multirow{4}{*}{60W} & \Th{AP$_{30:60}$} & 79.6 & 78.0 & 80.8 & 81.2 & 77.9 & 77.7 & 80.7 & 81.9\\
                                        & & \Th{AP$_{50}$} & 80.2 & 78.0 & 81.4 & 80.9 & 78.9 & 78.3 & 80.1 & 82.5 \\
                                        & & \Th{AR$_{50}$} & 89.3 & 85.8 & 89.6 & 88.6 & 86.3 & 86.5 & 88.9 & 91.0\\ \cmidrule{3-11}
                                      & & {\emph{t} (msec):} & 1.6 & 2.0 & 3.5 & 5.3 & 1.6 & 2.1 & 3.9 & 6.0 \\
\midrule
\multirow{4}{*}{AMD Versal SoC VCK190 (INT8) \tnote{1}} & \multirow{4}{*}{30W} & \Th{AP$_{30:60}$} & 57.6 & 55.8 & 50.5 & 53.1 & 65.7 & 54.5 & 59.8 & 63.1\\
                                        & & \Th{AP$_{50}$} & 57.6 & 51.7 & 45.3 & 51.6 & 64.6 & 50.9 & 58.0 & 62.6 \\
                                        & & \Th{AR$_{50}$} & 72.4 & 68.8 & 69.4 & 74.2 & 77.6 & 72.1 & 73.2 & 78.7\\ \cmidrule{3-11}
                                      & & {\emph{t} (msec):} & 0.24 & 0.37 & 0.67 & 1.0 & 0.47 & 0.83 & 0.83 & 1.34 \\                                    
\bottomrule
\end{tabular}
\vspace{-5pt}
}
\caption{\emph{Performance evaluation of YOLOv5 and YOLOv8 models across different devices.} \Th{AP}, {AR}: average precision, recall (\%) at different intersection over union (IoU) thresholds; \Th{Latency}: inference time (milliseconds) for a single image (batch=1); \Th{Power}: device power (Watts); $^1$half precision (FP16).}
\vspace{-10pt}
\label{tab:benchmark}
\end{table*}

We use YOLOv5~\cite{yolov5} and YOLOv8~\cite{yolov8}. YOLOv5~\cite{yolov5}, developed by ultralytics~\cite{ultralytics}, is an object detection model designed for real-time applications. Key features include: Cross stage partial network (CSPNet)~\cite{cspnet} backbone, Spatial pyramid pooling (SPP)~\cite{spp} layer,  Path aggregation network (PANet)~\cite{panet} head. YOLOv8~\cite{yolov8}, developed by ultralytics too, introduces further optimizations for enhanced performance and accuracy: anchor-free detection, efficient convolutional design and online image augmentation.



\subsection{Devices} 
\label{sec:devices}

We adapt and deploy models on the following embedded devices, suitable for on-board data processing in space:
\begin{itemize}[itemsep=1pt, topsep=2pt]
    \item NVIDIA Jetson Nano 4GB \cite{nvidia_jetson_nano}
    \item NVIDIA Jetson AGX Orin 64GB development kit \cite{nvidia_jetson_agx_orin}
    \item Google Coral Edge TPU 1GB development kit \cite{google_coral_edge_tpu}
    \item AMD Versal SoC VCK190 evaluation kit \cite{amd_fpga_soc_vck190}
\end{itemize}
For reference, we also include: 
\begin{itemize}[itemsep=1pt, topsep=2pt]
    \item ARM64 Cortex A78AE CPU \cite{nvidia_jetson_agx_orin}
    \item x86 Intel Xeon Gold 5218R CPU \cite{intel_xeon_gold_5218r}
    \item  NVIDIA V100 Tensor Core GPU \cite{nvidia_v100_tensor_core}
\end{itemize}

These devices encompass diverse processing architectures, from general-purpose processors and GPUs to reprogrammable gate arrays and application-specific tensor processors. The NVIDIA Jetson Nano and Jetson AGX Orin utilize ARM64 CPUs with integrated GPUs (128-core Maxwell for the Nano and 2048-core Ampere for the Orin) to deliver scalable AI performance. The Google Coral Edge TPU features a custom ASIC designed for high-speed, low-power AI inferencing, optimized for TensorFlow Lite models. The AMD Versal SoC combines scalar processors, high-density programmable logic (FPGA), and AI engines targeting vector-based algorithms for real-time DSP and AI computation.

\subsection{Dataset}
\label{sec:dataset}

\begin{figure}[t]
    \centering
    \includegraphics[width=0.7\linewidth]{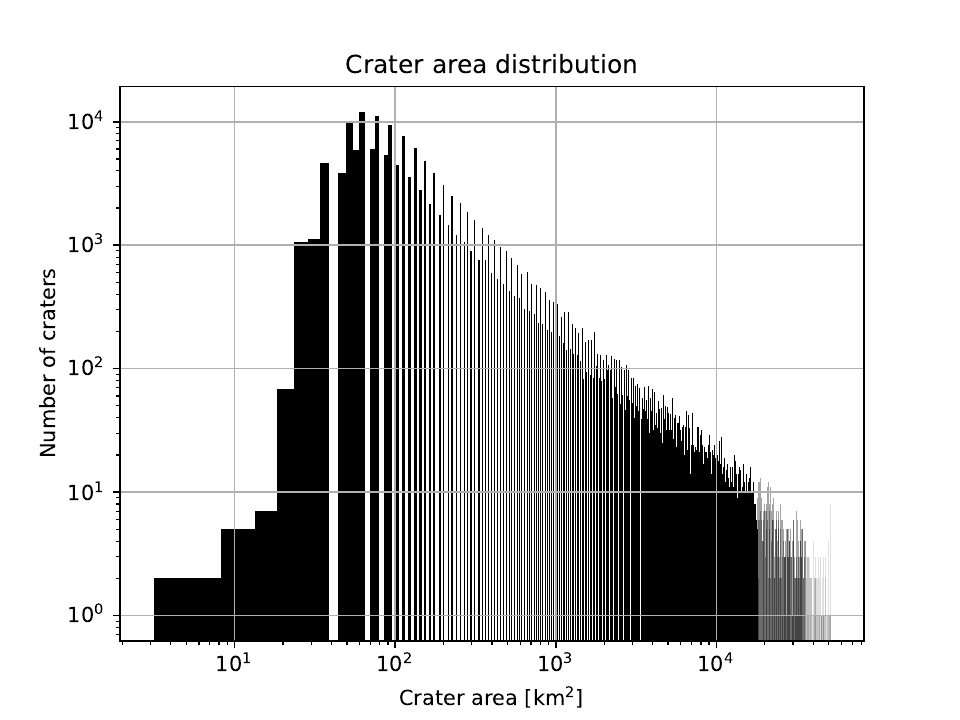}
    \caption{\emph{Crater Area Distribution} from Mars dataset~\cite{aichallenge}}
    \vspace{-10pt}
    \label{fig:crater_area}
\end{figure}

We employ a large-scale publicly available image dataset of Mars craters~\cite{aichallenge}, consisting of 102,675 images sourced from a global mosaic~\cite{THEMIS_2011}. These images are derived from the daytime infrared (DIR) data of the Mars Odyssey Thermal Emission Imaging System (THEMIS)~\cite{THEMIS_2004}, which covers the entire Martian surface with a resolution of 100 meters. Each image measures 25.6 by 25.6 km, translating to 256 x 256 pixels.
As illustrated in~\autoref{fig:crater_area}, most craters are small to medium in size, with the largest craters representing only a small fraction of the total. 

\subsection{Benchmark}
\label{sec:bench}

In \autoref{tab:benchmark}, we present benchmarking results for YOLOv5 and YOLOv8 models of various sizes, trained on the Mars craters dataset~\cite{aichallenge} using an NVIDIA V100 Tensor Core GPU. These models were evaluated across eight different devices, measuring Average Precision (AP) and Average Recall (AR) for detection accuracy, and inference time (t) for latency. Power consumption (in Watts) of each device was also recorded to provide insights into energy efficiency.

IoU thresholds are crucial for craters, which are generally small to medium-sized. Larger objects typically use IoU values above 0.5, indicating substantial overlap between predicted and ground truth bounding boxes. However, smaller objects like craters benefit from lower thresholds due to the higher impact of slight displacements in predictions. Given the predominance of small craters in the dataset (\autoref{fig:crater_area}), we use metrics with lower IoU thresholds. While maintaining standard AP$_{50}$ and AR$_{50}$ metrics, AP$_{30:60}$ is found to be more indicative compared to AP$_{50:90}$.


\subsection{Deployment}
\label{sec:deploy}

Deployment of the models on various architectures requires different approaches based on the underlying hardware and available vendor-specific tools.

For CPUs and the server GPU, deployment was straightforward using our trained PyTorch~\cite{pytorch} models without any specific adaptations.

Nvidia Jetson GPUs were employed by using TensorRT~\cite{tensorrt}. Models were converted from PyTorch to TensorRT engines via ONNX~\cite{onnx}, utilizing the YOLO export function provided by ultralytics, to achieve maximum performance.

For the Google Coral Edge TPU, PyTorch models were first converted to ONNX format, then to TensorFlow Lite~\cite{tflite} models, and finally optimized for the Edge TPU using the Edge TPU Compiler~\cite{coral_edgetpu_compiler}. This included quantization to INT8 precision for efficient inference.


Deploying on the Versal SoC VCK190 required significant model adjustments due to unsupported operations. We modified the forward function and post-processing procedures, replaced SiLU~\cite{silu} layers with LeakyReLU~\cite{leakyrelu}, and adjusted the DFL layer to remove unsupported functions. Using Vitis-AI~\cite{vitis_ai}, models were quantized to INT8 precision and compiled into DPU subgraphs for execution on Versal.

These varied approaches ensured efficient deployment across all platforms, leveraging each architecture's specific strengths.


\subsection{Results \& Discussion}
\label{sec:discuss}

Key factors influencing processing capabilities include the \emph{architecture} of the processing unit, enabling high levels of parallelization, and \emph{hardware efficiency}, enhancing the performance-to-power ratio. For instance, the Edge TPU, a specialized tensor processing unit for machine learning tasks, offers high efficiency and superior performance per watt. For a given architecture, increased power and smaller technology nodes typically improve performance.

\begin{figure}[h]
    \scriptsize
    \centering
    \begin{tikzpicture}[scale=0.7]
        \begin{axis}[
            font=\scriptsize,
            ybar,
            symbolic x coords={x86/ARM CPU, V100 GPU, Edge TPU, FPGA},
            xtick=data,
            ylabel={Average Precision (AP$_{50}$)},
            bar width=7pt,
            nodes near coords,
            nodes near coords style={font=\tiny},
            legend style={at={(0.5,-0.15)},anchor=north,legend columns=-1},
            enlarge x limits={abs=1cm},
            enlarge y limits={abs=0.3cm}
        ]
        \addplot coordinates {(x86/ARM CPU, 78.9) (V100 GPU, 78.9) (Edge TPU, 70.3) (FPGA, 65.7)};
        \addplot coordinates {(x86/ARM CPU, 78.3) (V100 GPU, 78.3) (Edge TPU, 72.9) (FPGA, 50.9)};
        \addplot coordinates {(x86/ARM CPU, 80.1) (V100 GPU, 80.2) (Edge TPU, 77.8) (FPGA, 58.0)};
        \addplot coordinates {(x86/ARM CPU, 82.5) (V100 GPU, 82.5) (Edge TPU, 78.5) (FPGA, 62.6)};
        \legend{nano, small, medium, large}
        \end{axis}
    \end{tikzpicture}
\caption{\emph{Crater detection AP$_{50}$} on Mars dataset~\cite{aichallenge} for distinct YOLOv8 versions across different architectures}.
    \vspace{-10pt}
    \label{fig:yolo_archs}
\end{figure}
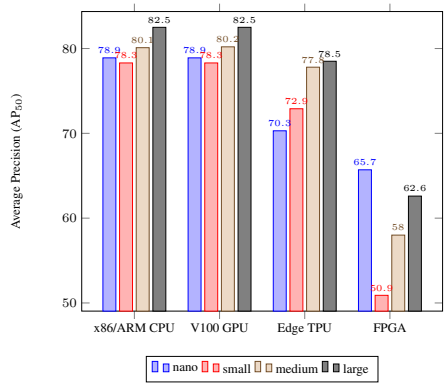

Device performance was quantified by measuring inference time for each model on all devices using single input images (batch size = 1). This approach can overlook other device capabilities, such as increased GPU occupancy/utilization, which could improve data throughput by processing multiple images simultaneously. Quantization on the Edge TPU and Versal SoC affected both recall and precision, converting model weights and activations from 32-bit floating-point numbers to 8-bit fixed-point integers, particularly in post-training quantization (\autoref{fig:yolo_archs}).

The Versal SoC showed the highest device performance but considerably lower model accuracy compared to other platforms, attributed to the extensive adjustments required for deployment; however, this can be ameliorated by retraining the model after alteration.Versal SoC achieved the highest processing performance but with relatively increased power consumption, which is a consideration for space avionics.

When including acceleration, the lowest power consumption capabilities were observed on TPU, providing the best performance/Watt in the 5–15W class of devices, comparable even to the Orin GPU built on a more advanced technology node. Edge TPU delivered an order of magnitude faster execution than CPU-based solutions, while the Orin GPU was even faster for larger models due to its larger memory. Overall, the 5–15W embedded devices show remarkable performance-accuracy-power benefits over conventional CPU-based avionics, whereas the 30–60W category enables handling 10–100x more demanding tasks.

\vspace{-5pt}
\section{Conclusions}
\label{sec:con}

We have introduced an approach to real-time crater detection on Mars using YOLO networks for embedded systems. We benchmarked these networks on a large Mars crater dataset across diverse systems. Our results highlight effective trade-offs between power consumption, computational efficiency and performance, demonstrating the feasibility of deploying deep learning networks for autonomous navigation, geological exploration or more. 
\vspace{5pt}
{\small 

\noindent \textbf{Acknowledgements}
Simon and Bill were supported by the RAMONES H2020 project (grant: 101017808).
}

\bibliographystyle{IEEEbib}
\bibliography{refs}

\end{document}